%% file: main.tex
\documentclass{article} 
\usepackage[preprint]{colm2025_conference}

\usepackage{microtype}
\usepackage{hyperref}
\usepackage{url}
\usepackage{booktabs}
\input{my_packages}

\usepackage{lineno}

\definecolor{darkblue}{rgb}{0, 0, 0.5}
\hypersetup{colorlinks=true, citecolor=darkblue, linkcolor=darkblue, urlcolor=darkblue}

\input{acronym}

\title{Adversarial Attacks on AI-Generated Text Detection Models:\\
A Token Probability-Based Approach Using Embeddings}

\input{authors}

\begin{document}

\ifcolmsubmission
\linenumbers
\fi

\maketitle
\input{sections/1-abstract}
\input{sections/2-introduction}

\input{sections/4-methodology}
\input{sections/8-result}
\input{sections/9-conclusion}
\input{sections/10-ethical_statment}
\bibliographystyle{colm2025_conference}
\bibliography{references}
\appendix
\input{sections/11-appendix}
\end{document}

%% file: my_packages.tex
\usepackage[nolist]{acronym}
\usepackage{verbatim}
\usepackage{float} 
\usepackage{multirow}
\usepackage{colortbl}
\usepackage{arydshln}
\usepackage{xcolor}

\usepackage{graphicx}
\usepackage{amsmath}

\usepackage{wrapfig}

%% file: acronym.tex
\begin{acronym}[DRAM]
    \acro{AI}{Artificial Intelligence}
    \acrodefindefinite{AI}{an}{an}
    \acro{NN}{Neural Network}
    \acro{DL}{Deep Learning}
    \acro{ML}{Machine Learning}
    \acro{TM}{Tsetlin Machine}
    \acro{NLP}{Natural Language Processing}
    \acro{LLMs}{Large Language Models}
    \acro{LLM}{Large Language Model}
    \acro{TM-AE}{Tsetlin Machine Auto-Encoder}
    \acro{RbE}{Reasoning by Elimination}
    \acro{TW}{Target Word}
    \acro{CoTM}{Coalesced Tsetlin Machine}
    \acrodefindefinite{ML}{an}{a}
    \acro{TA}{Tsetlin automaton}
    \acrodefplural{TA}{Tsetlin automata}
    \acro{TAT}{Tsetlin automaton team}
    \acro{GPT}{Generative Pre-trained Transformer}
    \acro{EDA}{Easy Data Augmentation}
    \acro{PLM}{Pre-Trained Language Model}
    \acro{PLMs}{Pre-Trained Language Models}
\end{acronym}

%% file: authors.tex
\author{Ahmed Khalid \& Lei Jiao \\
Department of ICT\\
University of Agder\\
Grimstad, Norway\\
\texttt{\{ahmed.k.kadhim, lei.jiao\}@uia.no} \\
\AND
Rishad Shafik\\
School of Engineering\\
Newcastle University\\
Newcastle upon Tyne, UK\\
\texttt{rishad.shafik@newcastle.ac.uk} \\
\AND
Ole-Christoffer Granmo\\
Department of ICT\\
University of Agder\\
Grimstad, Norway\\
\texttt{ole.granmo@uia.no}
}

%% file: sections/1-abstract.tex
\begin{abstract}
In recent years, text generation tools utilizing \ac{AI} have occasionally been misused across various domains, such as generating student reports or creative writings. 
This issue prompts plagiarism detection services to enhance their capabilities in identifying AI-generated content. 
Adversarial attacks are often used to test the robustness of AI-text generated detectors.
This work proposes a novel textual adversarial attack on the detection models such as Fast-DetectGPT. 
The method employs embedding models for data perturbation, aiming at reconstructing the \ac{AI} generated texts to reduce the likelihood of detection of the true origin of the texts. 
Specifically, we employ different embedding techniques, including the \ac{TM}, an interpretable approach in machine learning for this purpose.
By combining synonyms and embedding similarity vectors, we demonstrates the state-of-the-art reduction in detection scores against Fast-DetectGPT. Particularly, in the XSum dataset, the detection score decreased from 0.4431 to 0.2744 AUROC, and in the SQuAD dataset, it dropped from 0.5068 to 0.3532 AUROC.
\end{abstract}

%% file: sections/2-introduction.tex
\section{Introduction}

The responsibility of integrating into the scientific research and higher education community entails adhering to numerous behaviors and principles essential to safeguarding the integrity of educational and scientific progress. Consequently, the utilization of tools such as text editors or language enhancement applications must align with sound practices and uphold the ethical standards of scientific research and education~\citep{lund2023chatgpt,foltynek2023enai}. Despite the substantial advancements in \ac{AI} models, particularly within \ac{NLP} applications, there remains ongoing debate about the appropriate use of these tools for text generation \citep{leidner2017ethical,vsuster2017short}. This issue could significantly impact the integrity of the academic domain \citep{ENAI}. \ac{LLM} such as \ac{GPT}~\citep{gpt3,openai2022chatgpt,openai2023gpt4}, BERT~\citep{devlin2018bert}, and XLNet~\citep{xlnet} have gained widespread acceptance among users, demonstrating such high efficiency that distinguishing between human-generated and machine-generated content has become increasingly challenging~\citep{shahid2022you,ippolito2020automatic}.  

In response to this surge in the use of \ac{AI} techniques for text generation, various detection methods have been developed to ascertain the origin of text~\citep{solaiman2019release,fagni2021tweepfake,mitrovic2023chatgpt,gehrmann2019gltr,mitchell2023detectgpt,su2023detectllm}. These methods are categorized as supervised and unsupervised. Unsupervised methods are notable for their versatility in text detection, as they are capable of handling diverse text domains due to the nature of their pretraining ~\citep{gehrmann2019gltr,mitchell2023detectgpt}. These tools calculate probabilities and distribute them throughout the text. Under the zero-shot framework, it is presumed that AI-generated text exhibits a higher degree of probability variation compared to human-authored text, which can demonstrate both stability and variation. To assess such changes, additional text generation is required, significantly increasing execution time. The latest approach, FastDetect~\citep{fast-detect}, hypothesizes that AI-generated words are produced based on decisions made by the generative model, adhering to specific generation probabilities distinct from human word choices. For humans, word selection involves multiple influencing factors, rendering the output more personal and less statistically general. 

The majority of generated texts are fundamentally based on embedding principles in the initial stages of any \ac{LLMs}, where a dense vector in space is extracted for each word to encapsulate the contextual information present in the training dataset~\citep{goldberg2014word2vec,pennington2014glove}. The embedding maps words or tokens into a high-dimensional continuous space, which the model employs to facilitate various stages of text generation. This embedding, coupled with additional layers and algorithms used in \ac{LLMs}~\citep{attention}, informs the model’s predictions for subsequent words. Understanding this generative pattern enables the identification of AI-suggested content versus natural, human-like predictions, which are not strictly bound by patterns. 

Building on this foundation, in this paper, we propose an adversarial attack approach in which the embeddings are reverse engineered to exploit detection models by assigning low probability rates to predicted subsequent words, thereby lowering the overall text score as assessed by the detection systems.

Among the various embedding schemes considered in this work, we place particular emphasis on the \ac{TM}-based approach.  The \ac{TM} is an emerging machine learning technique that has demonstrated notable success in various applications \citep{tm,yadav2021human,tm-prf,parallel-tm,tm-edge}, including \ac{NLP} and computer vision. One architecture employed within the \ac{TM} framework is the \ac{TM-AE}, which is used in \ac{NLP} tasks to generate word embeddings~\citep{tm-ae}. The \ac{TM} is distinguished by its interpretability and the transparency of its output~\citep{tm,drop-clause,clause-size,yadav2021human}.  \ac{TM-AE} can produce word embeddings that encapsulate the contextual information conveyed by words derived from the training dataset~\citep{rbe_ahmed}. Consequently, \ac{TM} can be leveraged to modify texts under examination, using its contextual insights to influence AI-generated text evaluation tools, thereby creating an interpretable adversarial attack that compromises their detection capabilities.  

This work makes two primary innovations:  
\begin{enumerate}
    \item Proposing a novel adversarial attack on AI-origin detection systems, reducing detection accuracy from 0.4431 to 0.2744 AUROC on the XSum dataset and from 0.5068 to 0.3532 AUROC on the SQuAD dataset, through data perturbation with embedding models leveraging similar word probability vectors.
    \item Employing the interpretable TM model to gain deeper insights into the adversarial attack mechanism and its effects on text origin detection systems.
\end{enumerate}

%% file: sections/4-methodology.tex
\input{sections/5-detectors}
\input{sections/5-tm}

%% file: sections/5-detectors.tex
\section{AI-Text Detection Exploration}
\label{sec:detectors}
To understand the process of creating an adversarial attack on AI-text detection tools, it is essential first to comprehend the underlying mechanisms for detection. This section will explore advanced models like DetectGPT~\citep{mitchell2023detectgpt} and Fast-DetectGPT~\citep{fast-detect}.

In \textbf{Detect-GPT}, the scoring process relies on perturbations introduced by making minor modifications, such as replacing or masking specific words. These perturbations are typically generated using GPT-based models, such as T5. That method engages the model’s execution logic through a sequence of operations, generally involving 100 perturbations per input, with the model processing each input independently. As a result, that approach increases the execution time for detection compared to, e.g., Fast-DetectGPT. Additionally, the scoring procedure processes each input and its corresponding perturbations sequentially, further contributing to the computational burden. Summary of Detect-GPT Features:
\begin{itemize}
    \item Perturbation-Based: Generates perturbed versions of the input text by making slight alterations using a masked language model (e.g., T5).
    \item Probability Curvature: Compares the log-probabilities of the original text with those of its perturbations.
\end{itemize}

The \textbf{Fast-DetectGPT} method, in contrast, is designed to expedite the identification of text origins by utilizing conditional probabilities. This approach mitigates the additional complexities introduced by Detect-GPT for handling original texts. Instead of generating perturbations, the model creates alternatives—typically around 10,000—for each input token. The conditional likelihood function for each token is then evaluated, thereby eliminating the need for perturbations. This method obviates the sequential invocation of generation models, thereby achieving faster performance. Furthermore, the generation of 10,000 samples is streamlined, requiring only a single pass through the scorer. Discrepancy is determined by comparing the log-likelihood of the original token with the mean expected likelihood of the sampled alternatives.  
Discrepancy Formula:  
{\small
\[
\text{Discrepancy} = \frac{\text{Log-likelihood (original)} - \text{Mean (alternatives)}}{\sqrt{\text{Variance (alternatives)}}}.
\]
}

Summary of Fast-DetectGPT Features:
\begin{itemize}
    \item Sampling-Based: Generates alternative tokens for each position in the input text based on the model’s probability distribution.
    \item Single Forward Pass: Samples from a categorical distribution derived from the model’s logits to streamline execution.  
\end{itemize}

From the above summary of the AI-text detectors, it can be inferred that determining the origin of a text fundamentally relies on assessing the discrepancy between model-suggested probabilities and the natural probabilities characteristic of human decision-making. This principle is utilized in industrial tools, such as \textbf{Turnitin}, to evaluate text originality. According to their website~\citep{turnitin}, Turnitin employs an AI model grounded in the text-generation methodology of LLMs. The text is segmented into groups of five to ten sentences, with overlapping segments to analyze contextual inclusivity. Each segment is then processed by an AI checking model, which assigns a score between 0 and 1, indicating whether the text was human-written (0) or AI-generated (1). The overall average of these scores represents the final percentage of AI-generated content in the text.  

The Turnitin model relies on the GPT-3 framework for scoring. GPT-3 is trained on extensive Internet content, allowing it to generate text by predicting the most likely next words based on its training data. This prediction process is primarily governed by the transformer architecture~\citep{attention}, specifically its decoder, which determines the subsequent symbol during text generation. The key components of this mechanism are:
\begin{enumerate}
    \item Embedding Layer: Comprising two types—input embedding, which calculates contextual information for each symbol, and position embedding, which identifies each symbol’s placement within a sentence. 
    \item Transformer Layer: Includes the self-attention mechanism, which evaluates the importance of sentence components based on individual symbols. 
    \item Output Layer (Softmax): Processes the preceding results across the vocabulary to select the most likely next word.
\end{enumerate}

Based on the above observations, the sentence score can be lowered by replacing words with alternatives subject to the absolute probabilities within the embedding vector. This reduces the likelihood of selecting the next word, thereby decreasing the sentence’s overall score. As a result, the probability of detecting the text as AI-generated is reduced.

%% file: sections/5-tm.tex
\section{Proposed Adversarial Strategy}
In this section, 
we first introduce the framework for the adversarial attack using the embedding models and then we discuss the \ac{TM-AE} architecture and elaborate on its implementation within this study.

\subsection{Adversarial Framework}

\begin{figure}[ht]
    \centering
    \includegraphics[width=0.4\linewidth]{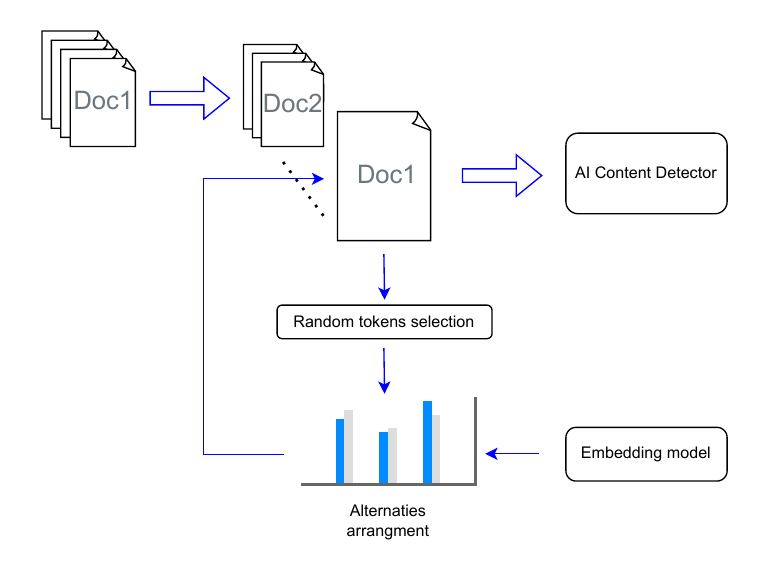}
    \caption{Proposed Adversarial Attack Framework. This figure illustrates the proposed design of an adversarial attack, where the input text (e.g., Doc1, Doc2, ...) is perturbed by selecting alternative tokens with low probability scores generated by embedding models, with the goal of misleading the detection model.}
    \label{fig:alternatives}
\end{figure}

We aim to develop an adversarial attack scheme on AI-generated text detection tools to deceive these models into classifying AI-generated texts as human-written. Embedding models are employed to guide the probability distribution during the selection of alternative words for replacement. Figure \ref{fig:alternatives} illustrates the proposed design of the adversarial attack scheme targeting AI-generated text detection systems. By replacing targeted words, the overall text score decreases compared to scores typically associated with AI-generated or human-written texts. Embedding models, which compute dense vectors for each token in the vocabulary, are integral to establishing relationships between tokens within a high-dimensional space. These vectors have been widely adopted in \ac{PLM} model architectures, particularly in Transformer-based models.

This work incorporates three methods for constructing probability distributions for tokens:
\subsubsection{Embedding Vector of Similarity}
The first method uses the original vector produced by the embedding model to identify alternatives based on their similarity to the target token. Cosine similarity is used as the similarity metric, and two primary parameters are introduced:  
\begin{itemize}
    \item \textbf{Similarity Vector Length}: Defines the length of the similarity vector derived from the embedding model’s original vector. In most cases the length is 400.
    \item \textbf{Similarity Threshold}: Specifies the minimum similarity score for selecting alternatives to the target token.
\end{itemize}


\subsubsection{Synonym Similarity Vector}
The second method utilizes grammatically correct synonyms sourced from external lexical databases, such as WordNet, which organizes English words into sets of cognitive synonyms crafted by human linguists and experts. The similarity between the target token and its synonyms is computed using the embedding vector and cosine similarity. Alternatives are ranked based on their similarity scores, with a predefined similarity degree used to select the final alternatives.


\subsubsection{Hybrid Scheme with Synonym and
Embedding Vectors}
{\color{black}The third method is a hybrid approach of the above two,  conducted in two stages. In the first stage, a synonym was randomly selected from WordNet. In the second stage, this synonym was replaced with a low-probability word derived from the knowledge vector generated by an embedding model. The \ac{TM-AE} embedding model was chosen for this scheme due to its transparent and interpretable structure.}

\subsection{Implementation  of Tsetlin Machine}
\label{sec:tm_ae}
The options among the deep-learning based approaches lack of interpretability due to their black-box nature. To address this, the \ac{TM} architecture was employed to enhance synonym substitution and assess the impact of these substitutions on text origin detection tools. The \ac{TM} provides interpretability by allowing detailed insights into decision-making processes, making it a suitable choice for understanding the implications of adversarial attacks. Below we explain the operational concept of TM and how it is adapted to this work.

\begin{figure}
    \centering
    \includegraphics[width=0.8\linewidth]{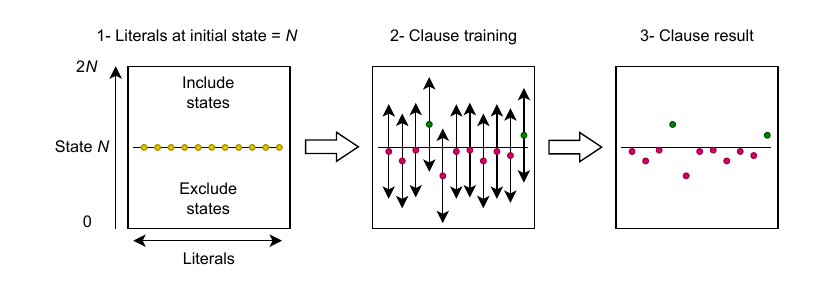}
    \caption{Clause formation in \ac{TM}. The $y$-axis is the state index while the $x$-axis is the literal index. When the state of an automaton is above $N$, its corresponding literal is included in the clause.  Before training, the states of the automata are configured as $N$ (the yellow dots in the left figure). During training, the states are updated (move up and down shown in the middle figure) based on the learning mechanism and training samples. Once trained, the clause is expressed by ANDing the included literals (the green dots in the right figure) and ignoring the excluded literals (red dots).  }
    \label{fig:tm}
\end{figure}

\ac{TM}~\citep{tm} effectively addresses complex pattern recognition tasks by leveraging propositional conjunctives, with each literal being managed by an individual automaton. Literals represent the entire vocabulary of the input, including their negations. For instance, in a sparse dataset where rows correspond to documents and columns correspond to literals (twice the size of the vocabulary), encoding assigns a value of 1 to a column if the corresponding $word$ exists in a document, while its negation receives the opposite value. If the $word$ does not appear in the document, the value is 0, and its negation is set to 1.

TM characterizes words through clauses, which are propositional expressions formed by included literals. For example, to describe the word \textit{car}, its clause could include the literals for words \textit{door}, \textit{wheels}, and \textit{no wing}, where \textit{door} and \textit{wheels} are literals in the original form, and \textit{no wing} arises from the literal representing the negation of the word \textit{wing}. Although \textit{wing} does not directly contribute to the clause, its negation plays a role in forming the description.

Clauses are iteratively trained to represent a word. In more detail, the final form of a clause ($C$) is represented as a conjunctive propositional expression consisting of literals that surpass the threshold state $N$ (See Figure \ref{fig:tm}). This clause, among others with a size of $n$ clauses with their selective literals, collectively define the output class ($y$). See Eq.~(\ref{eqn:output_function}).

\begin{equation}
    y = u\left(\sum_{j=1}^{n} C_j(X)\right).
\label{eqn:output_function}
\end{equation}
Here $u$ is a unit step thresholding, $u(v) = 1 ~\mathbf{if}~ v \ge 0 ~\mathbf{else}~ 0$, and $X$ is the input vector. Clearly, as seen from Eq. (\ref{eqn:output_function}), one word can have multiple corresponding clauses. For instance, the word \textit{heart} can have one clause with \textit{love} and \textit{woman}, and another clause with \textit{old} and \textit{hospital}.


The \ac{TM-AE} architecture~\citep{tm-ae} employed in this design is based on an enhanced \ac{TM} structure~\citep{cotm}, incorporating a coalesced weight matrix $W$ to facilitate the simultaneous training of multiple outputs and enabling nested elections among clauses. 
For example, for the word \textit{heart}, the weight assigned to the clause containing \textit{love} and \textit{woman} may differ from the weight assigned to the clause containing \textit{old} and \textit{hospital}, depending on the training context.
Particularly, in this work, the weights from \ac{TM-AE} are utilized to compute similarity probabilities for target words.

%% file: sections/8-result.tex
\section{Results}
Section~\ref{sec:data_model_spec} in the appendix includes further details that offer a concise overview of the datasets, \ac{PLMs}, detection models, and embedding models used.
In the experiments, the Area Under the Receiver Operating Characteristic (AUROC) score was utilized to evaluate the performance of AI-generated text detection models. An AUROC score of 1.0 signifies perfect detection, indicating the model's certainty that the text was AI-generated. Conversely, an AUROC score of 0.5 represents a random detection performance. Perturbations were generated for each dataset sample using PLM source models, as outlined in~\citet{fast-detect}, which served as the baseline source for text samples in this study. Subsequently, an additional perturbation was applied based on the proposed approach, as detailed below.

\subsection{Experiments with Embedding Vector of Similarity}
The datasets were sampled first using five PLM source models (GPT-2 XL, OPT-2.7, GPT-Neo-2.7, GPT-J-6, and GPT-NeoX-20) to generate AI-text samples, and then perturbed using six embedding models (GloVe, FastText, Word2Vec, TM-AE, ELMo, and BERT). The maximum permissible word change ratio was set to 5\% of each sample, with an average text length of approximately 150 words across all datasets (XSum, SQuAD, and Writing Prompts). Consequently, the number of altered words did not exceed eight, and in most cases, fewer words were changed. The process involved filtering each text to exclude non-informative tokens that do not represent valid English words. The remaining tokens were then checked for their presence in the vocabulary of the embedding model used. As the embedding models have limited vocabularies, not all words could be replaced.

\begin{table*}
\centering
\begin{tabular}{lrrrrrrrr}
\toprule
Source Model & Baseline   & BERT & ELMo & FastText & GloVe & TM-AE & Word2Vec \\
\midrule
\rowcolor[gray]{0.9} \multicolumn{8}{c}{The White-Box Environment} \\
 GPT-J-6     & 0.9866 & 0.7887 & 0.762  & 0.7270 & 0.7428 & 0.7268 & \textbf{0.7248} \\
 GPT-Neo-2.7 & 0.9946 & 0.8338 & 0.8056 & 0.7741 & 0.7815 & 0.7795 & \textbf{0.7700} \\
 GPT-NeoX-20 & 0.9744 & 0.7061 & 0.6677 & \textbf{0.6403} & 0.6612 & 0.6464 & 0.6483 \\
 GPT-2 XL    & 0.9953 & 0.8240 & 0.7988 & 0.7778 & \textbf{0.7620} & 0.7713 & 0.7779 \\
 OPT-2.7     & 0.9918 & 0.774  & 0.7379 & 0.7203 & 0.7292 & 0.7263 & \textbf{0.7137} \\
\hdashline
 Avg.         & 0.9885 & 0.7853 & 0.7544 & 0.7279 & 0.7353 & 0.7301 & \textbf{0.7269} \\
\hline

\rowcolor[gray]{0.9} \multicolumn{8}{c}{The Black-Box Environment} \\
 GPT-J-6     &0.9601 & 0.7538 & 0.7343 & 0.7068 &  0.7292 & \textbf{0.6926} & 0.7052 \\
 GPT-Neo-2.7 &0.9988 & 0.9399 & 0.9307 & 0.9187 &  0.9334 & \textbf{0.9163} & 0.9209 \\
 GPT-NeoX-20 &0.9412 & 0.6998 & 0.6686 & 0.6387 &  0.6773 & \textbf{0.6302} & 0.6485 \\
 GPT-2 XL    &0.9847 & 0.8541 & 0.8402 & 0.8184 &  0.8475 & \textbf{0.8109} & 0.8240 \\
 OPT-2.7     &0.9595 & 0.7854 & 0.7636 & \textbf{0.7356} &  0.7689 & 0.7387 & 0.7358 \\
\hdashline
 Avg.         &0.9689 & 0.8066 & 0.7875 & 0.7636 &  0.7913 & \textbf{0.7577} & 0.7669 \\
\bottomrule
\end{tabular}
\caption{Compare the detection performance (AUROC scores) of Fast-DetectGPT for various embedding models, evaluated in both white-box and black-box environments, highlighting the effect of embedding diversity on detection effectiveness.}
\label{fast-detect-vs-ds}
\end{table*}

Table \ref{fast-detect-vs-ds} illustrates the detection performance (AUROC scores) for all datasets (XSum, SQuAD, and Writing Prompts) using the Fast-DetectGPT. Two experimental scenarios were considered. 
\textbf{White-Box Environment}: Here, the text-generation source model is known, and the same model is used for scoring.
\textbf{Black-Box Environment}: A different model is used for scoring, specifically GPT-Neo-2.7.

\textbf{Results Overview}:
Word2Vec demonstrated the highest effectiveness for adversarial attacks, reducing the detection score to 0.7269 on average in the white-box scenario and 0.7669 in the black-box scenario.
BERT showed the least impact, with scores of 0.7853 and 0.8066 in the white-box and black-box environments, respectively, making it the most resistant embedding model.
TM-AE exhibited competitive performance, achieving the lowest average detection score of 0.7577 in the black-box environment and 0.7301 in the white-box environment, closely rivaling Word2Vec.
The results suggest that BERT's transformer-based architecture produces probability distributions similar to those of the PLMs used for text generation. This alignment makes it easier for detection models to identify AI-generated content, as evidenced by the relatively high scores. In contrast, embedding models like TM-AE and Word2Vec, which leverage distinct probability distributions, posed greater challenges to detection models.

\begin{wrapfigure}{r}{0.5\textwidth}
    \centering
    \includegraphics[width=1\linewidth]{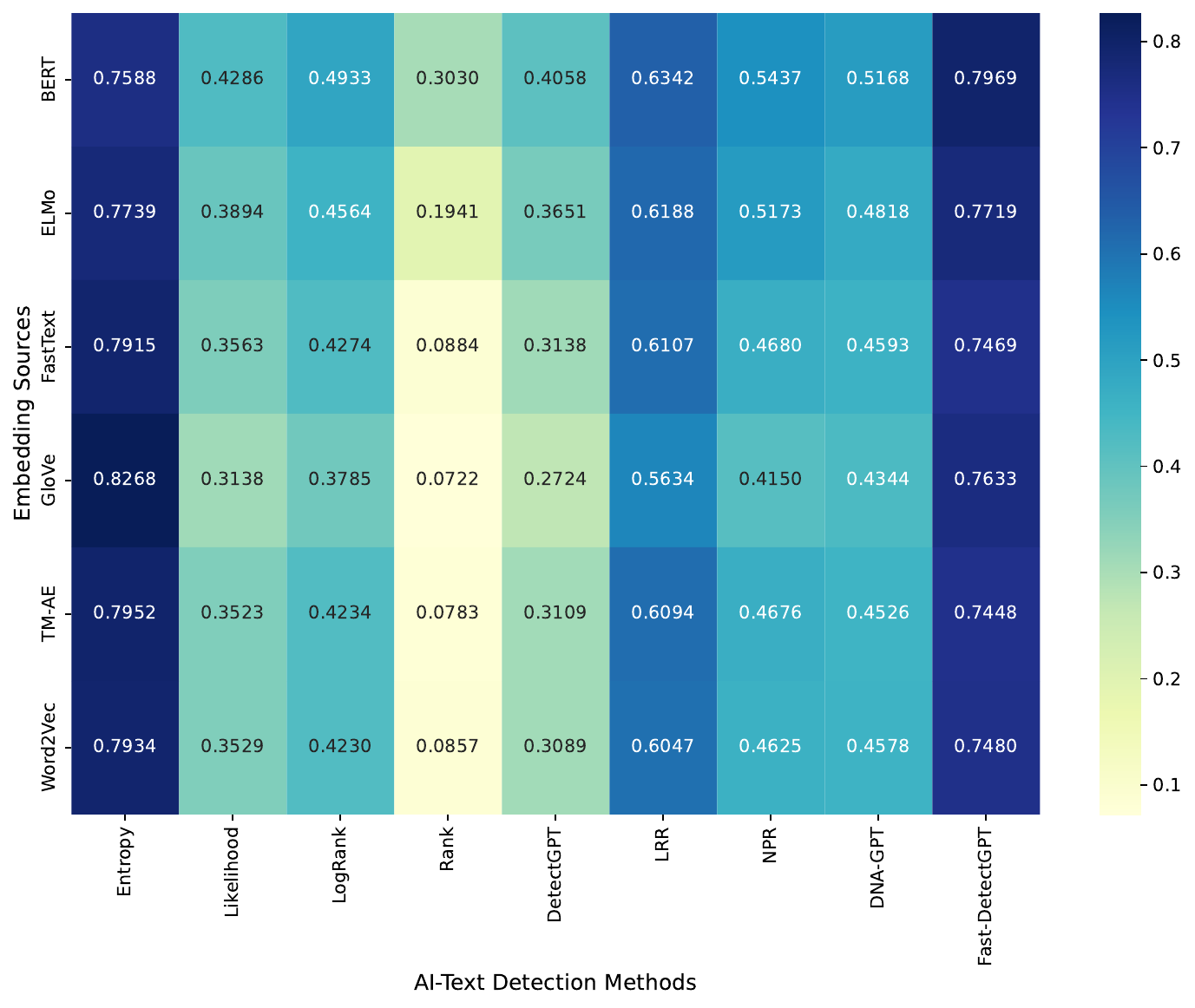}
    \caption{Heatmap illustrating the AUROC scores across various AI-text detection methods and embedding models. The x-axis represents the detection methods, while the y-axis corresponds to embedding models. 
    }
    \label{fig:heatmap-all}
\end{wrapfigure}

\textbf{Embedding Models Across Detection Methods}:
To further evaluate the effectiveness of embedding models, an additional experiment was conducted involving nine detection methods in the white-box environment (Fast-DetectGPT, Detect-GPT, NPR, LRR, DNA, Likelihood, Rank, LogRank, and Entropy) and four detection methods in the black-box environment (Fast-DetectGPT, Detect-GPT, NPR, LRR). Six embedding models (GloVe, FastText, Word2Vec, TM-AE, ELMo, and BERT) were tested across three datasets (XSum, SQuAD, and Writing Prompts), with five PLMs (GPT-2 XL, OPT-2.7, GPT-Neo-2.7, GPT-J-6, and GPT-NeoX-20).
{\color{black}In the experiments involving the detectors (Fast-DetectGPT, LRR, and NPR), the PLM model GPT-Neo-2.7 was used as the scoring model. Also, for the detectors LRR and NPR, perturbation generation employed the PLM model T5-3B.} Figure \ref{fig:heatmap-all} presents a heat map summarizing the results, with embedding models represented by rows and detection methods by columns.

\textbf{Key Findings}:
Fast-DetectGPT and Entropy were the most resilient detection methods, maintaining AUROC scores above 0.8 for several embedding models.
Rank performed poorly, with scores dropping below 0.1 for many embedding models.
GloVe achieved favorable results, effectively lowering detection performance across various methods.
BERT was least effective in representing adversarial attacks, consistently achieving high detection scores, corroborating earlier findings about its alignment with expected probability distributions.
TM-AE and Word2Vec delivered consistently strong performance across different detection methods, demonstrating their utility for adversarial attacks.
For detailed experimental results, refer to the tables provided in the appendix (Section~\ref{sec:all_other_exp}), which outline the performance of each method individually.

\subsection{Experiments Utilizing Synonym Similarity Vectors}
Previous experiments may not guarantee grammatically and semantically accurate substitutions to represent adversarial attacks, as they rely solely on embedding models and the dense vectors these models provide to generate replacements. In practical scenarios, constructing such adversarial representations would benefit from ensuring substitutions do not disrupt the context or alter the original meaning of the text. To achieve this, the following experiments employed human-curated synonym sets from WordNet to determine replacements for target words. For instance, the word car has synonyms such as $motorcar$, $railcar$, $auto$, $cable car$, $machine$, $elevator car$, $automobile$, $railroad car$, $railway car$, $gondola$. Replacing the target word with an appropriate synonym does not affect the overall sentence meaning. In these experiments, an embedding model was utilized to rank the likelihood of these synonyms, and the effect of varying both the number of substituted words and the nature of the substitution—whether the synonym was the most similar, intermediate, or least similar in the probability vector—was analyzed.

\begin{wrapfigure}{r}{0.5\textwidth}
  \centering
  \includegraphics[width=1\linewidth]{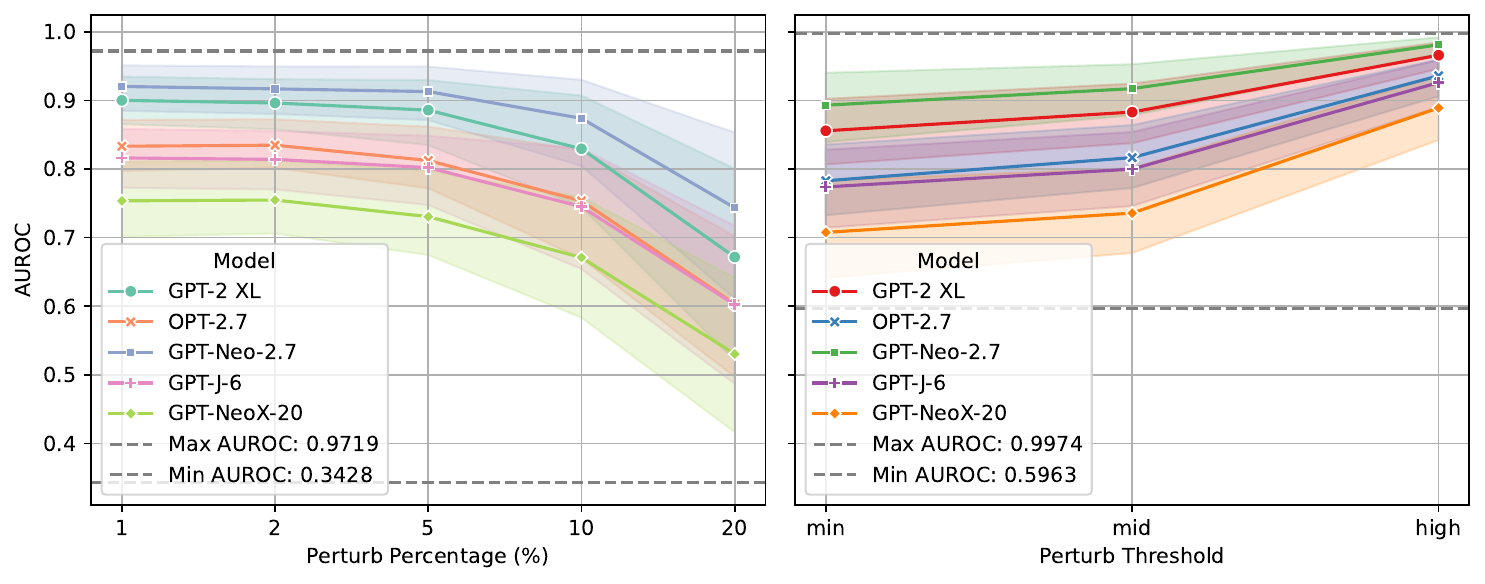}
  \caption{Impact of disturbance percentage and perturbation threshold on detection accuracy. The left panel shows the effect of increasing the percentage of replaced words, while the right panel illustrates the influence of synonym proximity (min, mid, high).
  }
  \label{fig:percentage}
\end{wrapfigure}

\textbf{Impact of Disturbance Percentage on Detection Accuracy}
Figure~\ref{fig:percentage} (left) illustrates the effect of increasing the percentage of word replacements (disturbance percentage) on the detection accuracy. As discussed previously, the percentage represents the maximum allowable substitutions rather than the exact number of replaced words. In this experiment, synonyms were selected from the middle of the synonym probability vector, with Fast-DetectGPT as the detection method and Word2Vec as the embedding model. The experiment was conducted under both white-box and black-box environments across three datasets (XSum, SQuAD, Writing Prompts) and five PLMs (GPT-2 XL, OPT-2.7, GPT-Neo-2.7, GPT-J-6, GPT-NeoX-20).

The results show a clear trend: as the disturbance percentage increased, the detection performance declined significantly. For example, the detection accuracy dropped from approximately 0.9 to 0.6 as the disturbance percentage increased from 1\% to 20\%. In certain cases, the detection score fell below 0.35 at a 20\% disturbance (Min AUROC is 0.3428), indicating a complete failure of the detection method to identify the origin of the text.


\textbf{Effect of Synonym Location on Detection Accuracy}
Figure~\ref{fig:percentage} (right) examines the impact of varying the position of the selected synonym in the probability vector on detection accuracy, using the same experimental settings as above. The min, mid, and high labels correspond to the least similar, intermediate, and most similar synonyms, respectively, based on the probability rankings. The results demonstrate that selecting lower-probability synonyms leads to a greater decline in detection accuracy. The detection scores decreased from over 0.95 to below 0.75 on average across detection models when low-probability synonyms were used, highlighting the vulnerability of detection methods to such substitutions.
\\

\subsection{Hybrid Experiments Using Synonym and Embedding Similarity Vectors}

In the hybrid model, the interpretability of the \ac{TM-AE} embedding is particularly advantageous. Since the first stage is human-understandable (craft by human), the transparency of the second stage, which uses \ac{TM-AE}, ensures that the entire two-stage process remains fully traceable. This transparency allows us to track completely the inference process of word replacement. For instance, for the target word \textit{car}, which is eventually replaced by $engine$, we can observe how the synonym $machine$ was selected in the first stage and then how $machine$ is replaced by $engine$ in the second stage. This human-understandable nature is critical for further analysis and debugging. 

In \ac{TM-AE}, the knowledge associated with a word is derived from the documents within the training database. For example, the target word \textit{car} might be represented by a clause containing \textit{engine} and \textit{not wing} after training. This representation can be based on a set of training documents where \textit{car} frequently appears alongside \textit{engine} but not with \textit{wing}. Notably, during the preparation of training data, word frequency is disregarded. The model considers only the presence or absence of a word (a Boolean value) when updating the corresponding column in the vocabulary of the input sparse vector. Essentially, the TM operates like an electoral system, where clauses vote on the target word using weights that are iteratively updated during training, ultimately forming a detailed and interpretable description of the word. Further details on the interpretability of the TM model can be found in works such as~\citet{tm-ae}, ~\citet{yadav2021human}, and~\citet{rbe}.

\begin{wrapfigure}{r}{0.5\textwidth}
  \centering
    \includegraphics[width=1\linewidth]{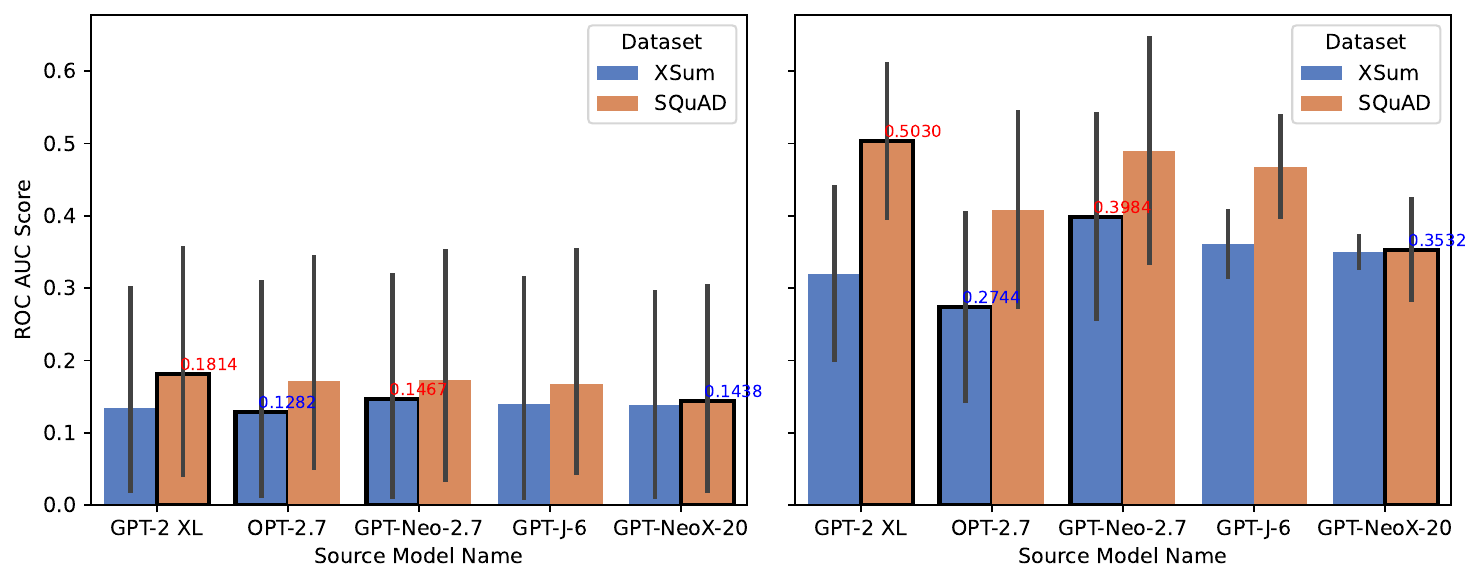}
    \caption{The impact of hybrid substitutions on detection accuracy across five PLMs, comparing all detection methods (left) and the Fast-DetectGPT method (right). Average detection scores for all methods fell to values between 0.1 and 0.2. For XSum, detection scores ranged from 0.1282 to 0.1467, while scores for SQuAD ranged from 0.1438 to 0.1814 (see blue and red numbers above dataset bars). The Fast-DetectGPT results (right) demonstrated average detection scores ranging between 0.2 and 0.5. Specifically, detection scores for XSum varied from 0.2744 to 0.3984, while scores for SQuAD ranged from 0.3532 to 0.5030.}
  \label{fig:tm-ae-roc}
\end{wrapfigure}


\textbf{Hybrid Substitution Results}
Figure \ref{fig:tm-ae-roc} shows the impact of hybrid substitutions on detection accuracy across five PLMs (GPT-2 XL, OPT-2.7, GPT-Neo-2.7, GPT-J-6, GPT-NeoX-20), comparing nine detection methods (left) and only the Fast-DetectGPT method (right). Detection accuracy was evaluated using two datasets, XSum (blue) and SQuAD (orange). 
For the left result, all nine detection methods were evaluated under the white-box environment (Fast-DetectGPT, Detect-GPT, NPR, LRR, DNA, Likelihood, Rank, LogRank, Entropy), and four methods under the black-box environment (Fast-DetectGPT, Detect-GPT, NPR, LRR).

The left results indicate a significant decline in detection accuracy, with scores falling to values between 0.1 and 0.2. Detection methods generally performed better on the SQuAD dataset compared to XSum. Moreover, as the complexity of the PLM source model increased—e.g., with GPT-NeoX-20, which contains more parameters and exhibits greater language generalization—the detection scores became more uniform, averaging below 0.15 across both datasets. This consistency suggests that highly generalized PLMs pose a greater challenge for detection methods, resulting in more uniform performance irrespective of the text's origin.

The Fast-DetectGPT results (right) show average detection scores fluctuating between 0.2 and 0.5. Specifically, detection scores for XSum varied from 0.2744 to 0.3984, while scores for SQuAD ranged from 0.3532 to 0.5030. Compared to similar work~\citep{token_ensemble}, these results demonstrate a substantial reduction in detection scores. For the XSum dataset, the detection score decreased from 0.4431 to 0.2744 (Percentage Reduction = 38.07\%), highlighting a notable decline in the accuracy of text origin detection. Similarly, for the SQuAD dataset, the score dropped from 0.5068 to 0.3532 (Percentage Reduction = 30.30\%), indicating the effectiveness of the proposed hybrid substitution method in successfully representing an adversarial attack.



%% file: sections/9-conclusion.tex
\section{Conclusion}
This study demonstrates the efficacy of adversarial attacks leveraging embedding-based substitutions to challenge AI-text detection methods. Results reveal that embedding models such as TM-AE and Word2Vec, as well as hybrid substitution methods, significantly reduce detection accuracy, with hybrid approaches achieving detection the lowest scores, particularly in black-box settings and with complex PLMs. Conversely, BERT showed to be the least effective for adversarial attacks due to its probability distributions aligning closely with those of PLM generation models, making it easier for detection systems to identify AI-generated content. These findings underscore the vulnerabilities of detection systems and highlight the potential of interpretable embedding approaches for crafting sophisticated adversarial attacks.

%% file: sections/10-ethical_statment.tex
\section*{Ethical Statement}
This work inherently involves ethical considerations, as it explores methods that could potentially bypass systems designed to detect the origin of text. Such systems often play a critical role in evaluating academic, professional, or creative works, and misuse of these methods could lead to ethical challenges by undermining trust and accountability. The primary aim of this research is to advance cybersecurity by examining the vulnerabilities of text detection systems and identifying potential adversarial strategies that could compromise their reliability. Specifically, this study highlights the role of embedding models in crafting adversarial attacks and emphasizes the need for detection systems to incorporate robust measures against such vulnerabilities. To address this, we propose the development of hybrid detection systems that integrate embedding model-based probability distributions with those from large language models, as suggested in \citet{token_ensemble}, to enhance their resilience against adversarial attacks.

%% file: sections/11-appendix.tex
\newpage
\section*{Appendix}
\input{sections/3-related_works}

\input{sections/12-experiment_settings}

\section{Detailed Experimental Results}
\label{sec:all_other_exp}
This appendix presents detailed tables corresponding to the heatmap plot (Figure~\ref{fig:heatmap-all}), summarizing the performance of each detection method and embedding model utilized in the experiments. The results include AUROC scores for various detection methods, including Fast-DetectGPT, Detect-GPT, NPR, LRR, DNA, Likelihood, Rank, LogRank, and Entropy, each provided in a separate table. The experiments were conducted across all datasets (XSum, SQuAD, and Writing Prompts), with text samples generated using different PLM models (GPT-2 XL, OPT-2.7B, GPT-Neo-2.7B, GPT-J-6B, and GPT-NeoX-20B) and perturbed using diverse embedding models (GloVe, FastText, Word2Vec, TM-AE, ELMo, and BERT).

\begin{table*}[ht]
\centering
\resizebox{\textwidth}{!}{%
\begin{tabular}{lllrrrrrr}
\toprule
\textbf{Environment} & \textbf{Dataset} & \textbf{Model} & \textbf{BERT} & \textbf{ELMo} & \textbf{FastText} & \textbf{GloVe} & \textbf{TM-AE} & \textbf{Word2Vec} \\
\midrule
\multirow{5}{*}{White} & \multirow{5}{*}{SQuAD} 
& GPT-2 XL & 0.6500 & 0.5900 & 0.5900 & 0.5400 & 0.5900 & 0.6000 \\
& & GPT-J-6 & 0.4800 & 0.4500 & 0.4300 & 0.3900 & 0.4000 & 0.4100 \\
& & GPT-Neo-2.7 & 0.5300 & 0.5000 & 0.4600 & 0.4400 & 0.4700 & 0.4600 \\
& & GPT-NeoX-20 & 0.3900 & 0.3300 & 0.3300 & 0.3100 & 0.3200 & 0.3300 \\
& & OPT-2.7 & 0.5800 & 0.5500 & 0.4900 & 0.4600 & 0.5000 & 0.4700 \\
\hline
\multirow{5}{*}{White} & \multirow{5}{*}{Writing Prompts} 
& GPT-2 XL & 0.6300 & 0.5900 & 0.5800 & 0.5400 & 0.5700 & 0.5800 \\
& & GPT-J-6 & 0.5900 & 0.5500 & 0.5100 & 0.4800 & 0.5000 & 0.5000 \\
& & GPT-Neo-2.7 & 0.6400 & 0.6000 & 0.5700 & 0.5300 & 0.5600 & 0.5700 \\
& & GPT-NeoX-20 & 0.5400 & 0.5000 & 0.4600 & 0.4400 & 0.4400 & 0.4800 \\
& & OPT-2.7 & 0.5300 & 0.5000 & 0.4800 & 0.4400 & 0.4600 & 0.4800 \\
\hline
\multirow{5}{*}{White} & \multirow{5}{*}{XSum} 
& GPT-2 XL & 0.4700 & 0.4600 & 0.4400 & 0.4200 & 0.4200 & 0.4400 \\
& & GPT-J-6 & 0.4100 & 0.3600 & 0.3600 & 0.3500 & 0.3500 & 0.3700 \\
& & GPT-Neo-2.7 & 0.4600 & 0.4500 & 0.4200 & 0.4100 & 0.4200 & 0.4100 \\
& & GPT-NeoX-20 & 0.3900 & 0.3600 & 0.3500 & 0.3400 & 0.3400 & 0.3400 \\
& & OPT-2.7 & 0.4900 & 0.4300 & 0.4300 & 0.4300 & 0.4400 & 0.4300 \\
\hline
\end{tabular}
}
\caption{Performance of the DNA-GPT detection method under white-box environment across different datasets (SQuAD, Writing Prompts, and XSum). Results are presented for various embedding models (BERT, ELMo, FastText, GloVe, TM-AE, and Word2Vec) and PLM sources (GPT-2 XL, GPT-J-6B, GPT-Neo-2.7B, GPT-NeoX-20B, and OPT-2.7B).}
\end{table*}

\begin{table*}[ht]
\centering
\resizebox{\textwidth}{!}{%
\begin{tabular}{lllrrrrrr}
\toprule
\textbf{Environment} & \textbf{Dataset} & \textbf{Model} & \textbf{BERT} & \textbf{ELMo} & \textbf{FastText} & \textbf{GloVe} & \textbf{TM-AE} & \textbf{Word2Vec} \\
\midrule
\multirow{5}{*}{White} & \multirow{5}{*}{SQuAD} 
& GPT-2 XL & 0.6600 & 0.5900 & 0.5400 & 0.4700 & 0.5200 & 0.5400 \\
& & GPT-J-6 & 0.3400 & 0.2900 & 0.2400 & 0.2100 & 0.2300 & 0.2500 \\
& & GPT-Neo-2.7 & 0.5100 & 0.4400 & 0.3900 & 0.3400 & 0.4100 & 0.3900 \\
& & GPT-NeoX-20 & 0.2400 & 0.2000 & 0.1900 & 0.1500 & 0.1700 & 0.1800 \\
& & OPT-2.7 & 0.5200 & 0.4600 & 0.3900 & 0.3500 & 0.4000 & 0.4000 \\
\hline
\multirow{5}{*}{White} & \multirow{5}{*}{Writing Prompts} 
& GPT-2 XL & 0.6600 & 0.6000 & 0.5000 & 0.4200 & 0.5000 & 0.4800 \\
& & GPT-J-6 & 0.4100 & 0.3700 & 0.2900 & 0.2300 & 0.2900 & 0.2800 \\
& & GPT-Neo-2.7 & 0.5800 & 0.5100 & 0.4100 & 0.3500 & 0.4100 & 0.4100 \\
& & GPT-NeoX-20 & 0.3500 & 0.3100 & 0.2400 & 0.2000 & 0.2300 & 0.2400 \\
& & OPT-2.7 & 0.4700 & 0.4100 & 0.3300 & 0.2800 & 0.3300 & 0.3300 \\
\hline
\multirow{5}{*}{White} & \multirow{5}{*}{XSum} 
& GPT-2 XL & 0.5900 & 0.5500 & 0.5200 & 0.4700 & 0.5100 & 0.5100 \\
& & GPT-J-6 & 0.4000 & 0.3700 & 0.3500 & 0.2900 & 0.3600 & 0.3200 \\
& & GPT-Neo-2.7 & 0.5700 & 0.5300 & 0.5100 & 0.4600 & 0.4800 & 0.4900 \\
& & GPT-NeoX-20 & 0.3400 & 0.3000 & 0.2800 & 0.2400 & 0.2700 & 0.2600 \\
& & OPT-2.7 & 0.4700 & 0.4200 & 0.3900 & 0.3600 & 0.4200 & 0.3900 \\
\hline
\multirow{5}{*}{Black} & \multirow{5}{*}{SQuAD} 
& GPT-2 XL & 0.2200 & 0.1900 & 0.1600 & 0.1300 & 0.1500 & 0.1500 \\
& & GPT-J-6 & 0.2600 & 0.2300 & 0.1800 & 0.1600 & 0.1800 & 0.1900 \\
& & GPT-Neo-2.7 & 0.5100 & 0.4400 & 0.3900 & 0.3400 & 0.4100 & 0.3900 \\
& & GPT-NeoX-20 & 0.2400 & 0.2100 & 0.1800 & 0.1500 & 0.1600 & 0.1800 \\
& & OPT-2.7 & 0.2200 & 0.1900 & 0.1500 & 0.1200 & 0.1500 & 0.1500 \\
\hline
\multirow{5}{*}{Black} & \multirow{5}{*}{Writing Prompts} 
& GPT-2 XL & 0.3500 & 0.3200 & 0.2300 & 0.2000 & 0.2400 & 0.2300 \\
& & GPT-J-6 & 0.3500 & 0.3100 & 0.2500 & 0.2000 & 0.2400 & 0.2400 \\
& & GPT-Neo-2.7 & 0.5800 & 0.5100 & 0.4100 & 0.3500 & 0.4100 & 0.4100 \\
& & GPT-NeoX-20 & 0.3300 & 0.3000 & 0.2200 & 0.1900 & 0.2200 & 0.2300 \\
& & OPT-2.7 & 0.3200 & 0.2900 & 0.2100 & 0.1800 & 0.2000 & 0.2100 \\
\hline
\multirow{5}{*}{Black} & \multirow{5}{*}{XSum} 
& GPT-2 XL & 0.3200 & 0.3000 & 0.2700 & 0.2500 & 0.2500 & 0.2600 \\
& & GPT-J-6 & 0.2700 & 0.2600 & 0.2300 & 0.1900 & 0.2300 & 0.2200 \\
& & GPT-Neo-2.7 & 0.5700 & 0.5300 & 0.5100 & 0.4600 & 0.4800 & 0.4900 \\
& & GPT-NeoX-20 & 0.2700 & 0.2500 & 0.2200 & 0.2000 & 0.2200 & 0.2200 \\
& & OPT-2.7 & 0.3000 & 0.2500 & 0.2400 & 0.2100 & 0.2500 & 0.2300 \\
\hline
\end{tabular}
}
\caption{Performance of the Detect-GPT detection method under white-box and black-box environments across different datasets (SQuAD, Writing Prompts, and XSum). Results are provided for various embedding models (BERT, ELMo, FastText, GloVe, TM-AE, and Word2Vec) and PLM sources (GPT-2 XL, GPT-J-6B, GPT-Neo-2.7B, GPT-NeoX-20B, and OPT-2.7B).}
\end{table*}

\begin{table*}[ht]
\centering
\resizebox{\textwidth}{!}{%
\begin{tabular}{lllrrrrrr}
\toprule
\textbf{Environment} & \textbf{Dataset} & \textbf{Model} & \textbf{BERT} & \textbf{ELMo} & \textbf{FastText} & \textbf{GloVe} & \textbf{TM-AE} & \textbf{Word2Vec} \\
\midrule
\multirow{5}{*}{White} & \multirow{5}{*}{SQuAD} 
& GPT-2 XL & 0.7500 & 0.7700 & 0.7800 & 0.8100 & 0.7900 & 0.7800 \\
& & GPT-J-6 & 0.7800 & 0.7900 & 0.8200 & 0.8400 & 0.8100 & 0.8200 \\
& & GPT-Neo-2.7 & 0.7800 & 0.7900 & 0.8100 & 0.8400 & 0.8000 & 0.8100 \\
& & GPT-NeoX-20 & 0.8000 & 0.8200 & 0.8300 & 0.8600 & 0.8500 & 0.8400 \\
& & OPT-2.7 & 0.7400 & 0.7400 & 0.7800 & 0.8100 & 0.7800 & 0.7800 \\
\hline
\multirow{5}{*}{White} & \multirow{5}{*}{Writing Prompts} 
& GPT-2 XL & 0.7100 & 0.7300 & 0.7600 & 0.8100 & 0.7500 & 0.7600 \\
& & GPT-J-6 & 0.7300 & 0.7500 & 0.7700 & 0.8300 & 0.7800 & 0.7700 \\
& & GPT-Neo-2.7 & 0.6700 & 0.7000 & 0.7200 & 0.7700 & 0.7300 & 0.7200 \\
& & GPT-NeoX-20 & 0.7600 & 0.7800 & 0.7900 & 0.8400 & 0.8000 & 0.8000 \\
& & OPT-2.7 & 0.7100 & 0.7300 & 0.7600 & 0.8100 & 0.7600 & 0.7500 \\
\hline
\multirow{5}{*}{White} & \multirow{5}{*}{XSum} 
& GPT-2 XL & 0.7600 & 0.7700 & 0.7800 & 0.8000 & 0.7800 & 0.7800 \\
& & GPT-J-6 & 0.8400 & 0.8500 & 0.8600 & 0.8900 & 0.8600 & 0.8600 \\
& & GPT-Neo-2.7 & 0.8000 & 0.8100 & 0.8200 & 0.8500 & 0.8300 & 0.8300 \\
& & GPT-NeoX-20 & 0.8200 & 0.8400 & 0.8500 & 0.8700 & 0.8500 & 0.8500 \\
& & OPT-2.7 & 0.7300 & 0.7400 & 0.7500 & 0.7900 & 0.7600 & 0.7600 \\
\hline
\end{tabular}
}
\caption{Performance of the Entropy detection method under white-box environment across different datasets (SQuAD, Writing Prompts, and XSum). Results are presented for various embedding models (BERT, ELMo, FastText, GloVe, TM-AE, and Word2Vec) and PLM sources (GPT-2 XL, GPT-J-6B, GPT-Neo-2.7B, GPT-NeoX-20B, and OPT-2.7B).}
\end{table*}

\begin{table*}[ht]
\centering
\resizebox{\textwidth}{!}{%
\begin{tabular}{lllrrrrrr}
\toprule
\textbf{Environment} & \textbf{Dataset} & \textbf{Model} & \textbf{BERT} & \textbf{ELMo} & \textbf{FastText} & \textbf{GloVe} & \textbf{TM-AE} & \textbf{Word2Vec} \\
\midrule
\multirow{5}{*}{White} & \multirow{5}{*}{SQuAD} 
& GPT-2 XL & 0.9200 & 0.8800 & 0.8800 & 0.8600 & 0.8800 & 0.8800 \\
& & GPT-J-6 & 0.7700 & 0.7400 & 0.7100 & 0.7200 & 0.7000 & 0.7000 \\
& & GPT-Neo-2.7 & 0.8400 & 0.8100 & 0.7800 & 0.7800 & 0.7800 & 0.7700 \\
& & GPT-NeoX-20 & 0.6100 & 0.5600 & 0.5500 & 0.5500 & 0.5500 & 0.5500 \\
& & OPT-2.7 & 0.8600 & 0.8200 & 0.8000 & 0.7900 & 0.8000 & 0.7900 \\
\hline
\multirow{5}{*}{White} & \multirow{5}{*}{Writing Prompts} 
& GPT-2 XL & 0.8900 & 0.8700 & 0.8500 & 0.8500 & 0.8500 & 0.8500 \\
& & GPT-J-6 & 0.8900 & 0.8700 & 0.8300 & 0.8500 & 0.8300 & 0.8300 \\
& & GPT-Neo-2.7 & 0.8900 & 0.8600 & 0.8300 & 0.8400 & 0.8300 & 0.8300 \\
& & GPT-NeoX-20 & 0.8500 & 0.8200 & 0.7900 & 0.8200 & 0.7900 & 0.8000 \\
& & OPT-2.7 & 0.7600 & 0.7200 & 0.7100 & 0.7200 & 0.6900 & 0.6900 \\
\hline
\multirow{5}{*}{White} & \multirow{5}{*}{XSum} 
& GPT-2 XL & 0.7400 & 0.7200 & 0.6900 & 0.6700 & 0.6800 & 0.6900 \\
& & GPT-J-6 & 0.7100 & 0.6800 & 0.6400 & 0.6600 & 0.6500 & 0.6400 \\
& & GPT-Neo-2.7 & 0.7700 & 0.7500 & 0.7100 & 0.7200 & 0.7300 & 0.7100 \\
& & GPT-NeoX-20 & 0.6600 & 0.6200 & 0.5800 & 0.6200 & 0.6000 & 0.5900 \\
& & OPT-2.7 & 0.7000 & 0.6700 & 0.6500 & 0.6700 & 0.6800 & 0.6600 \\
\hline
\multirow{5}{*}{Black} & \multirow{5}{*}{SQuAD} 
& GPT-2 XL & 0.8900 & 0.8700 & 0.8600 & 0.8700 & 0.8600 & 0.8500 \\
& & GPT-J-6 & 0.7600 & 0.7300 & 0.7100 & 0.7200 & 0.6800 & 0.7100 \\
& & GPT-Neo-2.7 & 0.9500 & 0.9400 & 0.9300 & 0.9400 & 0.9300 & 0.9400 \\
& & GPT-NeoX-20 & 0.6700 & 0.6300 & 0.6200 & 0.6400 & 0.6000 & 0.6200 \\
& & OPT-2.7 & 0.8300 & 0.8100 & 0.7800 & 0.8000 & 0.7800 & 0.7800 \\
\hline
\multirow{5}{*}{Black} & \multirow{5}{*}{Writing Prompts} 
& GPT-2 XL & 0.9100 & 0.9000 & 0.8800 & 0.9100 & 0.8800 & 0.8900 \\
& & GPT-J-6 & 0.8600 & 0.8200 & 0.8000 & 0.8300 & 0.7900 & 0.8000 \\
& & GPT-Neo-2.7 & 0.9600 & 0.9500 & 0.9400 & 0.9500 & 0.9300 & 0.9400 \\
& & GPT-NeoX-20 & 0.8200 & 0.8000 & 0.7600 & 0.8000 & 0.7500 & 0.7700 \\
& & OPT-2.7 & 0.8300 & 0.8000 & 0.7800 & 0.8200 & 0.7700 & 0.7700 \\
\hline
\multirow{5}{*}{Black} & \multirow{5}{*}{XSum} 
& GPT-2 XL & 0.7700 & 0.7500 & 0.7200 & 0.7600 & 0.7000 & 0.7300 \\
& & GPT-J-6 & 0.6500 & 0.6500 & 0.6100 & 0.6300 & 0.6100 & 0.6100 \\
& & GPT-Neo-2.7 & 0.9100 & 0.9100 & 0.8900 & 0.9100 & 0.8900 & 0.8900 \\
& & GPT-NeoX-20 & 0.6100 & 0.5800 & 0.5400 & 0.5900 & 0.5500 & 0.5500 \\
& & OPT-2.7 & 0.7000 & 0.6800 & 0.6500 & 0.6800 & 0.6700 & 0.6600 \\
\hline
\end{tabular}
}
\caption{Performance of the Fast-DetectGPT detection method under white-box and black-box environments across different datasets (SQuAD, Writing Prompts, and XSum). Results are provided for various embedding models (BERT, ELMo, FastText, GloVe, TM-AE, and Word2Vec) and PLM sources (GPT-2 XL, GPT-J-6B, GPT-Neo-2.7B, GPT-NeoX-20B, and OPT-2.7B).}
\end{table*}

\begin{table*}[ht]
\centering
\resizebox{\textwidth}{!}{%
\begin{tabular}{lllrrrrrr}
\toprule
\textbf{Environment} & \textbf{Dataset} & \textbf{Model} & \textbf{BERT} & \textbf{ELMo} & \textbf{FastText} & \textbf{GloVe} & \textbf{TM-AE} & \textbf{Word2Vec} \\
\midrule
\multirow{5}{*}{White} & \multirow{5}{*}{SQuAD} 
& GPT-2 XL & 0.8100 & 0.7900 & 0.8100 & 0.7500 & 0.7800 & 0.7900 \\
& & GPT-J-6 & 0.6500 & 0.6300 & 0.6100 & 0.5800 & 0.6100 & 0.6100 \\
& & GPT-Neo-2.7 & 0.7300 & 0.7000 & 0.7000 & 0.6600 & 0.7100 & 0.6900 \\
& & GPT-NeoX-20 & 0.5300 & 0.5000 & 0.5300 & 0.4600 & 0.5000 & 0.5000 \\
& & OPT-2.7 & 0.7400 & 0.7400 & 0.7000 & 0.6800 & 0.7200 & 0.7100 \\
\hline
\multirow{5}{*}{White} & \multirow{5}{*}{Writing Prompts} 
& GPT-2 XL & 0.7800 & 0.7600 & 0.7500 & 0.6900 & 0.7500 & 0.7500 \\
& & GPT-J-6 & 0.7300 & 0.7100 & 0.7000 & 0.6400 & 0.7000 & 0.6900 \\
& & GPT-Neo-2.7 & 0.7800 & 0.7600 & 0.7400 & 0.6800 & 0.7300 & 0.7400 \\
& & GPT-NeoX-20 & 0.6800 & 0.6600 & 0.6400 & 0.6000 & 0.6400 & 0.6500 \\
& & OPT-2.7 & 0.7000 & 0.6700 & 0.6700 & 0.6100 & 0.6700 & 0.6800 \\
\hline
\multirow{5}{*}{White} & \multirow{5}{*}{XSum} 
& GPT-2 XL & 0.6600 & 0.6400 & 0.6500 & 0.6100 & 0.6400 & 0.6500 \\
& & GPT-J-6 & 0.5600 & 0.5600 & 0.5400 & 0.5100 & 0.5500 & 0.5400 \\
& & GPT-Neo-2.7 & 0.6500 & 0.6500 & 0.6500 & 0.5900 & 0.6500 & 0.6300 \\
& & GPT-NeoX-20 & 0.5300 & 0.5200 & 0.5100 & 0.4700 & 0.5100 & 0.5000 \\
& & OPT-2.7 & 0.6200 & 0.6000 & 0.5900 & 0.5500 & 0.6100 & 0.5800 \\
\hline
\multirow{5}{*}{Black} & \multirow{5}{*}{SQuAD} 
& GPT-2 XL & 0.5700 & 0.5600 & 0.5700 & 0.5100 & 0.5400 & 0.5500 \\
& & GPT-J-6 & 0.5800 & 0.5700 & 0.5400 & 0.5000 & 0.5300 & 0.5400 \\
& & GPT-Neo-2.7 & 0.7300 & 0.7000 & 0.7000 & 0.6600 & 0.7100 & 0.6900 \\
& & GPT-NeoX-20 & 0.4800 & 0.4600 & 0.4700 & 0.4000 & 0.4500 & 0.4500 \\
& & OPT-2.7 & 0.5700 & 0.5700 & 0.5400 & 0.5200 & 0.5600 & 0.5500 \\
\hline
\multirow{5}{*}{Black} & \multirow{5}{*}{Writing Prompts} 
& GPT-2 XL & 0.6500 & 0.6300 & 0.6000 & 0.5500 & 0.6100 & 0.6000 \\
& & GPT-J-6 & 0.6800 & 0.6700 & 0.6500 & 0.6000 & 0.6600 & 0.6500 \\
& & GPT-Neo-2.7 & 0.7800 & 0.7600 & 0.7400 & 0.6800 & 0.7300 & 0.7400 \\
& & GPT-NeoX-20 & 0.6100 & 0.6000 & 0.5800 & 0.5300 & 0.5800 & 0.5800 \\
& & OPT-2.7 & 0.6100 & 0.5900 & 0.5800 & 0.5200 & 0.5800 & 0.5900 \\
\hline
\multirow{5}{*}{Black} & \multirow{5}{*}{XSum} 
& GPT-2 XL & 0.4900 & 0.4700 & 0.4800 & 0.4500 & 0.4600 & 0.4600 \\
& & GPT-J-6 & 0.4900 & 0.4800 & 0.4700 & 0.4400 & 0.4800 & 0.4700 \\
& & GPT-Neo-2.7 & 0.6500 & 0.6500 & 0.6500 & 0.5900 & 0.6500 & 0.6300 \\
& & GPT-NeoX-20 & 0.4600 & 0.4500 & 0.4500 & 0.4100 & 0.4500 & 0.4400 \\
& & OPT-2.7 & 0.5300 & 0.5100 & 0.5100 & 0.4700 & 0.5200 & 0.5000 \\
\hline
\end{tabular}
}
\caption{Performance of the LRR detection method under white-box and black-box environments across different datasets (SQuAD, Writing Prompts, and XSum). Results are provided for various embedding models (BERT, ELMo, FastText, GloVe, TM-AE, and Word2Vec) and PLM sources (GPT-2 XL, GPT-J-6B, GPT-Neo-2.7B, GPT-NeoX-20B, and OPT-2.7B).}
\end{table*}

\begin{table*}[ht]
\centering
\resizebox{\textwidth}{!}{%
\begin{tabular}{lllrrrrrr}
\toprule
\textbf{Environment} & \textbf{Dataset} & \textbf{Model} & \textbf{BERT} & \textbf{ELMo} & \textbf{FastText} & \textbf{GloVe} & \textbf{TM-AE} & \textbf{Word2Vec} \\
\midrule
\multirow{5}{*}{White} & \multirow{5}{*}{SQuAD} 
& GPT-2 XL & 0.5400 & 0.4900 & 0.4700 & 0.4100 & 0.4500 & 0.4700 \\
& & GPT-J-6 & 0.3700 & 0.3300 & 0.2900 & 0.2600 & 0.2900 & 0.2900 \\
& & GPT-Neo-2.7 & 0.4300 & 0.3900 & 0.3500 & 0.3100 & 0.3600 & 0.3500 \\
& & GPT-NeoX-20 & 0.2600 & 0.2200 & 0.2000 & 0.1700 & 0.1900 & 0.2000 \\
& & OPT-2.7 & 0.4900 & 0.4500 & 0.4000 & 0.3600 & 0.4000 & 0.4000 \\
\hline
\multirow{5}{*}{White} & \multirow{5}{*}{Writing Prompts} 
& GPT-2 XL & 0.5900 & 0.5400 & 0.5000 & 0.4300 & 0.5000 & 0.5000 \\
& & GPT-J-6 & 0.5300 & 0.4800 & 0.4400 & 0.3700 & 0.4200 & 0.4200 \\
& & GPT-Neo-2.7 & 0.5800 & 0.5300 & 0.4900 & 0.4400 & 0.4800 & 0.4900 \\
& & GPT-NeoX-20 & 0.4600 & 0.4100 & 0.3700 & 0.3300 & 0.3600 & 0.3700 \\
& & OPT-2.7 & 0.4700 & 0.4300 & 0.4000 & 0.3400 & 0.3800 & 0.3900 \\
\hline
\multirow{5}{*}{White} & \multirow{5}{*}{XSum} 
& GPT-2 XL & 0.3900 & 0.3600 & 0.3300 & 0.3000 & 0.3200 & 0.3300 \\
& & GPT-J-6 & 0.2800 & 0.2600 & 0.2300 & 0.2000 & 0.2300 & 0.2300 \\
& & GPT-Neo-2.7 & 0.3700 & 0.3400 & 0.3200 & 0.2800 & 0.3000 & 0.3100 \\
& & GPT-NeoX-20 & 0.2700 & 0.2400 & 0.2100 & 0.1900 & 0.2200 & 0.2100 \\
& & OPT-2.7 & 0.4000 & 0.3600 & 0.3500 & 0.3100 & 0.3600 & 0.3400 \\
\hline
\end{tabular}
}
\caption{Performance of the Likelihood detection method under white-box environment across different datasets (SQuAD, Writing Prompts, and XSum). Results are presented for various embedding models (BERT, ELMo, FastText, GloVe, TM-AE, and Word2Vec) and PLM sources (GPT-2 XL, GPT-J-6B, GPT-Neo-2.7B, GPT-NeoX-20B, and OPT-2.7B).}
\end{table*}

\begin{table*}[ht]
\centering
\resizebox{\textwidth}{!}{%
\begin{tabular}{lllrrrrrr}
\toprule
\textbf{Environment} & \textbf{Dataset} & \textbf{Model} & \textbf{BERT} & \textbf{ELMo} & \textbf{FastText} & \textbf{GloVe} & \textbf{TM-AE} & \textbf{Word2Vec} \\
\midrule
\multirow{5}{*}{White} & \multirow{5}{*}{SQuAD} 
& GPT-2 XL & 0.6300 & 0.5800 & 0.5700 & 0.5000 & 0.5400 & 0.5600 \\
& & GPT-J-6 & 0.4300 & 0.3900 & 0.3600 & 0.3200 & 0.3600 & 0.3600 \\
& & GPT-Neo-2.7 & 0.5100 & 0.4700 & 0.4300 & 0.3800 & 0.4500 & 0.4300 \\
& & GPT-NeoX-20 & 0.3100 & 0.2700 & 0.2600 & 0.2200 & 0.2400 & 0.2500 \\
& & OPT-2.7 & 0.5600 & 0.5300 & 0.4700 & 0.4400 & 0.4900 & 0.4800 \\
\hline
\multirow{5}{*}{White} & \multirow{5}{*}{Writing Prompts} 
& GPT-2 XL & 0.6500 & 0.6100 & 0.5700 & 0.5000 & 0.5800 & 0.5700 \\
& & GPT-J-6 & 0.5900 & 0.5500 & 0.5100 & 0.4400 & 0.5000 & 0.5000 \\
& & GPT-Neo-2.7 & 0.6500 & 0.6000 & 0.5600 & 0.5000 & 0.5500 & 0.5600 \\
& & GPT-NeoX-20 & 0.5200 & 0.4700 & 0.4400 & 0.3900 & 0.4300 & 0.4400 \\
& & OPT-2.7 & 0.5400 & 0.4900 & 0.4700 & 0.4000 & 0.4600 & 0.4600 \\
\hline
\multirow{5}{*}{White} & \multirow{5}{*}{XSum} 
& GPT-2 XL & 0.4600 & 0.4300 & 0.4100 & 0.3700 & 0.4000 & 0.4100 \\
& & GPT-J-6 & 0.3400 & 0.3200 & 0.2900 & 0.2600 & 0.3000 & 0.2900 \\
& & GPT-Neo-2.7 & 0.4400 & 0.4200 & 0.4000 & 0.3500 & 0.3800 & 0.3800 \\
& & GPT-NeoX-20 & 0.3200 & 0.2900 & 0.2700 & 0.2400 & 0.2700 & 0.2600 \\
& & OPT-2.7 & 0.4600 & 0.4200 & 0.4000 & 0.3600 & 0.4100 & 0.4000 \\
\hline
\end{tabular}
}
\caption{Performance of the LogRank detection method under white-box environment across different datasets (SQuAD, Writing Prompts, and XSum). Results are presented for various embedding models (BERT, ELMo, FastText, GloVe, TM-AE, and Word2Vec) and PLM sources (GPT-2 XL, GPT-J-6B, GPT-Neo-2.7B, GPT-NeoX-20B, and OPT-2.7B).}
\end{table*}

\begin{table*}[ht]
\centering
\resizebox{\textwidth}{!}{%
\begin{tabular}{lllrrrrrr}
\toprule
\textbf{Environment} & \textbf{Dataset} & \textbf{Model} & \textbf{BERT} & \textbf{ELMo} & \textbf{FastText} & \textbf{GloVe} & \textbf{TM-AE} & \textbf{Word2Vec} \\
\midrule
\multirow{5}{*}{White} & \multirow{5}{*}{SQuAD} 
& GPT-2 XL & 0.8100 & 0.7800 & 0.7500 & 0.6700 & 0.7200 & 0.7300 \\
& & GPT-J-6 & 0.5300 & 0.4700 & 0.4300 & 0.3900 & 0.4200 & 0.4400 \\
& & GPT-Neo-2.7 & 0.6800 & 0.6500 & 0.6000 & 0.5400 & 0.6200 & 0.6000 \\
& & GPT-NeoX-20 & 0.3900 & 0.3600 & 0.3500 & 0.2900 & 0.3200 & 0.3300 \\
& & OPT-2.7 & 0.6800 & 0.6600 & 0.5800 & 0.5400 & 0.6200 & 0.6000 \\
\hline
\multirow{5}{*}{White} & \multirow{5}{*}{Writing Prompts} 
& GPT-2 XL & 0.7700 & 0.7400 & 0.6400 & 0.5500 & 0.6400 & 0.6200 \\
& & GPT-J-6 & 0.5800 & 0.5600 & 0.4700 & 0.3900 & 0.4800 & 0.4500 \\
& & GPT-Neo-2.7 & 0.7100 & 0.6600 & 0.5700 & 0.4900 & 0.5700 & 0.5700 \\
& & GPT-NeoX-20 & 0.5200 & 0.5000 & 0.4100 & 0.3700 & 0.4100 & 0.4300 \\
& & OPT-2.7 & 0.6000 & 0.5600 & 0.4900 & 0.4200 & 0.4800 & 0.4900 \\
\hline
\multirow{5}{*}{White} & \multirow{5}{*}{XSum} 
& GPT-2 XL & 0.7500 & 0.7100 & 0.7100 & 0.6600 & 0.6900 & 0.7100 \\
& & GPT-J-6 & 0.5700 & 0.5700 & 0.5400 & 0.4800 & 0.5600 & 0.5200 \\
& & GPT-Neo-2.7 & 0.7300 & 0.7200 & 0.7000 & 0.6500 & 0.7000 & 0.6800 \\
& & GPT-NeoX-20 & 0.5200 & 0.5100 & 0.4700 & 0.4400 & 0.4700 & 0.4600 \\
& & OPT-2.7 & 0.6400 & 0.6000 & 0.5800 & 0.5400 & 0.6000 & 0.5700 \\
\hline
\multirow{5}{*}{Black} & \multirow{5}{*}{SQuAD} 
& GPT-2 XL & 0.3300 & 0.3100 & 0.2900 & 0.2400 & 0.2600 & 0.2700 \\
& & GPT-J-6 & 0.3800 & 0.3500 & 0.2900 & 0.2700 & 0.2900 & 0.3000 \\
& & GPT-Neo-2.7 & 0.6800 & 0.6500 & 0.6000 & 0.5400 & 0.6200 & 0.6000 \\
& & GPT-NeoX-20 & 0.3500 & 0.3200 & 0.3000 & 0.2300 & 0.2700 & 0.2800 \\
& & OPT-2.7 & 0.3300 & 0.3100 & 0.2700 & 0.2200 & 0.2700 & 0.2700 \\
\hline
\multirow{5}{*}{Black} & \multirow{5}{*}{Writing Prompts} 
& GPT-2 XL & 0.4500 & 0.4300 & 0.3300 & 0.2800 & 0.3400 & 0.3300 \\
& & GPT-J-6 & 0.4500 & 0.4200 & 0.3500 & 0.2900 & 0.3500 & 0.3400 \\
& & GPT-Neo-2.7 & 0.7100 & 0.6600 & 0.5700 & 0.4900 & 0.5700 & 0.5700 \\
& & GPT-NeoX-20 & 0.4000 & 0.4000 & 0.3000 & 0.2600 & 0.3000 & 0.3100 \\
& & OPT-2.7 & 0.4000 & 0.3800 & 0.3000 & 0.2600 & 0.3000 & 0.3100 \\
\hline
\multirow{5}{*}{Black} & \multirow{5}{*}{XSum} 
& GPT-2 XL & 0.4400 & 0.4200 & 0.4100 & 0.3700 & 0.3800 & 0.3900 \\
& & GPT-J-6 & 0.3800 & 0.3700 & 0.3700 & 0.3000 & 0.3600 & 0.3400 \\
& & GPT-Neo-2.7 & 0.7300 & 0.7200 & 0.7000 & 0.6500 & 0.7000 & 0.6800 \\
& & GPT-NeoX-20 & 0.3700 & 0.3600 & 0.3300 & 0.3000 & 0.3300 & 0.3300 \\
& & OPT-2.7 & 0.4100 & 0.3700 & 0.3600 & 0.3300 & 0.3700 & 0.3500 \\
\hline
\end{tabular}
}
\caption{Performance of the NPR detection method under white-box and black-box environments across different datasets (SQuAD, Writing Prompts, and XSum). Results are provided for various embedding models (BERT, ELMo, FastText, GloVe, TM-AE, and Word2Vec) and PLM sources (GPT-2 XL, GPT-J-6B, GPT-Neo-2.7B, GPT-NeoX-20B, and OPT-2.7B).}
\end{table*}

\begin{table}[t]
\centering
\resizebox{\textwidth}{!}{%
\begin{tabular}{lllrrrrrr}
\toprule
\textbf{Environment} & \textbf{Dataset} & \textbf{Model} & \textbf{BERT} & \textbf{ELMo} & \textbf{FastText} & \textbf{GloVe} & \textbf{TM-AE} & \textbf{Word2Vec} \\
\midrule
\multirow{5}{*}{White} & \multirow{5}{*}{SQuAD} 
& GPT-2 XL & 0.3700 & 0.2400 & 0.1000 & 0.0700 & 0.0800 & 0.0900 \\
& & GPT-J-6 & 0.2200 & 0.1400 & 0.0400 & 0.0400 & 0.0400 & 0.0500 \\
& & GPT-Neo-2.7 & 0.2700 & 0.1900 & 0.0500 & 0.0400 & 0.0500 & 0.0700 \\
& & GPT-NeoX-20 & 0.1800 & 0.1100 & 0.0500 & 0.0200 & 0.0200 & 0.0400 \\
& & OPT-2.7 & 0.2800 & 0.2100 & 0.0600 & 0.0500 & 0.0500 & 0.0600 \\
\hline
\multirow{5}{*}{White} & \multirow{5}{*}{Writing Prompts} 
& GPT-2 XL & 0.4400 & 0.2500 & 0.1200 & 0.1000 & 0.1400 & 0.1200 \\
& & GPT-J-6 & 0.3800 & 0.2200 & 0.1200 & 0.1000 & 0.0900 & 0.1200 \\
& & GPT-Neo-2.7 & 0.4400 & 0.2700 & 0.1600 & 0.1200 & 0.1400 & 0.1300 \\
& & GPT-NeoX-20 & 0.3600 & 0.2000 & 0.1200 & 0.1100 & 0.1000 & 0.1200 \\
& & OPT-2.7 & 0.4200 & 0.2700 & 0.1300 & 0.1200 & 0.1200 & 0.1200 \\
\hline
\multirow{5}{*}{White} & \multirow{5}{*}{XSum} 
& GPT-2 XL & 0.2800 & 0.2000 & 0.0900 & 0.0700 & 0.0800 & 0.0800 \\
& & GPT-J-6 & 0.2100 & 0.1500 & 0.0600 & 0.0500 & 0.0700 & 0.0600 \\
& & GPT-Neo-2.7 & 0.2600 & 0.1700 & 0.0800 & 0.0700 & 0.0700 & 0.0800 \\
& & GPT-NeoX-20 & 0.2000 & 0.1400 & 0.0700 & 0.0600 & 0.0500 & 0.0600 \\
& & OPT-2.7 & 0.2400 & 0.1600 & 0.0600 & 0.0500 & 0.0700 & 0.0700 \\
\hline
\end{tabular}
}
\caption{Performance of the Rank detection method under white-box environment across different datasets (SQuAD, Writing Prompts, and XSum). Results are presented for various embedding models (BERT, ELMo, FastText, GloVe, TM-AE, and Word2Vec) and PLM sources (GPT-2 XL, GPT-J-6B, GPT-Neo-2.7B, GPT-NeoX-20B, and OPT-2.7B).}
\end{table}

%% file: sections/3-related_works.tex
\section{Related Work}
In related studies focusing on constructing adversarial attacks on AI-generated text detection systems, two primary approaches are identified in  \citet{token_ensemble} and in \citet{reliably_detected}, respectively. The approach~\citet{token_ensemble} involves replacing specific tokens in the text with words generated randomly by \ac{LLMs}. Multiple variants of GPT models are employed to generate a probability distribution for replacement, thereby influencing the source of the scoring. However, that method significantly increases the complexity of the attack due to the inherent opacity of \ac{LLMs}, which operate as black-box models, making it challenging to rationalize token selection. Furthermore, the reliance on large-scale \ac{LLMs} results in extensive execution times. 
Additionally, \citet{token_ensemble} employs the same models utilized for scoring or generating perturbations, thereby restricting the exposure of detection models to alternative types of probability distributions. This limitation arises because these detection models are inherently familiar with the probability distributions of the tokens they analyze. 
Conversely, approach ~\citet{reliably_detected} proposes relying on a single \ac{LLM} as the source for token replacement, thereby creating a paraphrase-based attack. This approach is more constrained than the former, as it significantly limits the diversity of probability distributions considered.

%% file: sections/12-experiment_settings.tex
\section{Experimental Resources}
\label{sec:data_model_spec}

A variety of datasets, \ac{PLMs}, AI-text detection models, and embedding models were utilized in this study. Except for the embedding models, the remaining datasets and models align with those used in \citet{fast-detect}, aiming to evaluate the proposed approach comprehensively. This diversity ensures a broad evaluation scope, encompassing various scenarios across downstream applications. The following subsections provide a concise overview of the datasets, \ac{PLMs}, detection models, and embedding models employed.

\subsection{Datasets}
Three English datasets from diverse domains were used in the experiments. The XSum dataset, introduced by~\citet{xsum_ds}, is designed for abstractive summarization and provides 500 concise summaries of news articles. The SQuAD dataset~\citep{squad_ds}, based on Wikipedia contexts, includes 300 samples for training models to answer questions. The Writing Prompts dataset~\citep{writing_ds} contains 500 samples aimed at generating creative and coherent stories based on prompts. These datasets address distinct yet overlapping text generation tasks, enhancing various natural language processing capabilities.


\subsection{Pre-trained Language Models (PLMs)}
Several pre-trained language models (PLMs) with transformer-based architectures were employed during different stages of preparation and scoring,  both during the pre-detection sample collection phase and the text perturbation process for detection. The PLMs utilized include GPT-2 XL with approximately 1.5 billion parameters, OPT-2.7B and GPT-Neo-2.7B with 2.7 billion parameters each, T5-3B with 3 billion parameters, GPT-J-6B with 6 billion parameters, and GPT-NeoX-20B with 20 billion parameters.


\subsection{AI-Text Detection Models}
The study utilized a set of classifiers designed for zero-shot evaluation of adversarial attacks, including: \textbf{Fast-Detect} and \textbf{Detect-GPT}: As detailed in Section~\ref{sec:detectors}.
     \textbf{Normalized Perturbation Rank (NPR)} and \textbf{Log Probability and Log Rank (LRR)}: Both leverage rank- and probability-based features to enhance accuracy while balancing computational efficiency~\citep{su2023detectllm}.  
     \textbf{LogRank}, \textbf{Likelihood}, and \textbf{Rank}: Metrics based on token probabilities and their ranks to evaluate the likelihood of text being AI-generated~\citep{gehrmann2019gltr,solaiman2019release}.  
    \textbf{Entropy} and \textbf{Divergent N-Gram Analysis (DNA)}: Techniques focusing on distributional irregularities and n-gram variations to detect machine-generated text~\citep{ippolito2020automatic,yang2023dna}.


\subsection{Embedding Models}
A diverse set of embedding models was employed to calculate token probability distributions, ensuring variety in the representation of text. The embedding models used include: 
     \textbf{GloVe}: Pre-trained embeddings capturing word relationships based on co-occurrence statistics.
     \textbf{FastText}: Embeddings incorporating character-level information, effectively handling out-of-vocabulary words.
     \textbf{Word2Vec}: Static embeddings generated using CBOW or Skip-Gram methods for context prediction.
    \textbf{TM-AE}: As detailed in Section~\ref{sec:tm_ae}, this model uses logical expressions for word embeddings.
    \textbf{ELMo}: Contextual embeddings derived from bidirectional language models (BiLSTMs).
     \textbf{BERT}: Contextual embeddings utilizing Transformers, capturing bidirectional context.
These models were trained on the One Billion Word dataset~\citep{one_billion}, with a vocabulary size limited to 40,000 tokens.
